# Gait Identification under Surveillance Environment based on Human Skeleton

Xingkai Zheng    Xirui Li    Ke Xu*    Xinghao Jiang    Tanfeng Sun

School of Cyber Science and Engineering, Shanghai Jiao Tong University, Shanghai, China

(jdxccz, lixirui142, 113025816, xhjiang, tfsun) @sjtu.edu.cn

*Abstract*—As an emerging biological identification technology, vision-based gait identification is an important research content in biometrics. Most existing gait identification methods extract features from gait videos and identify a probe sample by a query in the gallery. However, video data contains redundant information and can be easily influenced by bagging (BG) and clothing (CL). Since human body skeletons convey essential information about human gaits, a skeleton-based gait identification network is proposed in our project. First, extract skeleton sequences from the video and map them into a gait graph. Then a feature extraction network based on Spatio-Temporal Graph Convolutional Network (ST-GCN) is constructed to learn gait representations. Finally, the probe sample is identified by matching with the most similar piece in the gallery. We tested our method on the CASIA-B dataset. The result shows that our approach is highly adaptive and gets the advanced result in BG, CL conditions, and average.

## I. INTRODUCTION

Like the face, fingerprint, and iris, gait is also a biometric characteristic, which can be used for personal identity authentication. Compared with other biometrics, gait recognition can be carried out directly through video information without the help of additional instruments and can be recognized from a certain distance. At the same time, the cost of a camouflage gait is higher than other features. Therefore, gait recognition is very suitable as an independent or auxiliary personal authentication method for fugitive arrest, forensic identification, and social security.

There are some problems with the existing vision-based gait recognition. First of all, gait is easily affected by conditions such as clothing and carrying, and it is easy to judge a person's gait as others in different states. Secondly, limited by the surveillance camera, we can only obtain the picture in a single direction but also lose the three-dimensional information, which increases the difficulty of recognition. In addition, the gait model is easy to overfit on the test set. These problems increase the difficulty of gait recognition. At present, there are three main gait recognition methods.

Appearance-based gait recognition processes the gait sequence as an image. For example, the gait energy map [1] aligns and averages the human contour in a gait cycle, and then the features can be extracted by the depth learning method. Therefore, this method is very dependent on the accuracy and alignment of the contour, and it is easy to be affected by clothing and the environment.

Bone-based gait recognition uses the pose and movement of joint points to describe gait. Bones can be extracted from images using keypoint estimation [2] or motion capture sensors. Therefore, the bone-based method is very dependent on the positioning accuracy of joint points. Different clothes and angles will affect the extraction of joint points, and some occlusion can easily cause the loss of joint points.

The method based on data fusion combines different data types as recognition methods, such as optical flow information and depth information [3]. The problem with this method is that it is challenging to process multimodal data in one architecture. At the same time, integrating different data brings additional computational overhead, which limits the application of this method.

Most of the existing methods use appearance-based gait recognition. Although the image contains rich information, most of the content of gait recognition belongs to redundant data, which will increase the cost of recognition and be very sensitive to clothing and the environment. At the same time, the joint point extraction methods based on deep learning are gradually improved, such as OpenPose and AlphaPose, which have high efficiency and accuracy. The bone information extracted by these methods is equivalent to removing a large amount of unnecessary, redundant information for gait recognition and retaining the critical bone point information. It not only reduces the computational overhead but also increases the information density.

Therefore, in this subject, to improve the effectiveness of the bone-based method, we propose a gait feature extraction network based on ST-GCN [4]. For the extracted bone points, we connect them into a map according to the human body and then combine the corresponding bone points of different frames to form a bone sequence diagram containing spatial and temporal information. After that, ST-GCN is input to continuously fuse the information of points in a small range similar to image convolution. Finally, the gait features are obtained.

## II. RELATED WORKS

### A. Geometry-based bone descriptors

Geometry-based bone descriptors can be used for behavior recognition. Some works use three-dimensional bone descriptors [5], and others integrate different bone geometric features [6]. In the field of gait bone description, many studies further extract depth features from geometric features. There is work to use CNN + LSTM to learn depth features from joint coordinates [7]. There is work using RNN architecture to learn features from calculated geometric features [8]. There is also work to stack bones as skiGEI and extract features with 3d-CNN [9].

### B. GCN-based bone descriptors

The template is used to format your paper and style the text. All margins, column widths, line spaces, and text fonts are prescribed; please do not alter them. You may note peculiarities. For example, the head margin in this template measures proportionately more than is customary. This measurement and others are deliberate, using specifications that anticipate your paper as one part of the entire proceedings and not as an independent document. Please do not revise any of the current designations.

### III. APPROACH

The system collects a gait sequence from the video in units of frames, extracts the bone points to form a Spatio-temporal map, and puts it into the graph convolutional network for comparison and recognition. The overall structure of the proposed method is shown in Fig. 1.

### A. Problem definition

Suppose there are N people, K samples, the i-th person has the identity, the j-th sample is $x_j\ j \in \{1,2,...,K\}$, and there is a mapping $X \rightarrow Y$, that is, each sample x exists in its corresponding identity y. The K-segment samples are divided into a library sample and a sample to be tested. The identity $y_{gallery}$ of the library sample $x_{gallery}$ is known, and the identity of the sample $x_{probe}$ to be tested is unknown. We hope to use some methods to predict the identity of the sample to be tested.

We define it as a matching problem, that is, for each sequence to be tested, search for matching items in all library sequences and predict its label based on this. This method can be expressed as

$$I_p = \{y_i | i = argmax_{i \in \{1,2,...n\}} s_i\}$$
$$s_i = D(f(x_p), f(x_i))$$

where $f(\cdot)$ extracts the characteristics of the gait sequence, and $D(\cdot)$ is the matching function. The L2 paradigm is usually used to calculate the matching degree, such as $D(a,b) = ||a-b||$. The recognition result $I_p$ is the identity $y_i$ of the sequence $x_p$ with the largest matching degree between 1 to n of all sequences in the library.

### B. Gait skeleton diagram

We use the gait skeleton diagram to represent each sample x. First, we use OpenPose, an open-source human body two-dimensional pose estimation application, to extract the skeleton sequence from the original gait video. For the extracted skeleton sequence, suppose there are T frames, and N joint points are extracted in each frame. We can build a spatiotemporal graph of bones $G = (V, E)$. Where $V = \{v_{ti} | t = 1,2,...,T, i = 1,2,...,N\}$. The upper edge set in the spatial dimension is the same as the joint connection of the human body. Assume that in the human body, the neighbor of the point $v_{ti}$ is $N(v_{ti})$, and in the time dimension, it connects the corresponding points of adjacent frames, which is

$$E = \{(v_{ti}, v_{tN(v_{ti})}) | t = 1,2...,T, i = 1,2,...,N\} + \{(v_{ti}, v_{t'i}) | t = 1,2,...,T, i = 1,2,...,N, t' \in \{t+1, t-1\})\}$$

### C. ST-GCN

Spatial Temporal Graph Convolutional Network (ST-GCN) imitates the convolution operation performed on the picture. For image data such as the human skeleton sequence, the convolution operation can be performed from the two dimensions of time and space to extract its features. In this project, ST-GCN is used as the basis to extract the features of the gait sequence. First, you need to define the range of the convolution kernel when performing convolution operations, that is, define the size of a point neighborhood $B(v)$:

$$B(v_{ti}) = \{v_{qj} | d(v_{tj}, v_{ti}) \leq K, |q-t| \leq \left\lfloor \frac{\Gamma}{2} \right\rfloor \}$$

where for the neighborhood set $B(v_{ti})$ of the point $v_{ti}$, $K$ controls the range in the space dimension, and $\Gamma$ controls the range in the time dimension. $d(\cdot)$ represents the path distance between two points in the figure. Therefore, we define a point within the distance $K$ in space and within the range of $\left\lfloor \frac{\Gamma}{2} \right\rfloor$ in time as its neighbors.

After defining the neighborhood, in order to assign weights to different types of points in the convolution operation, it is necessary to assign labels to the points in the neighborhood to distinguish them. Many different grouping methods have been proposed before, and we use a method that takes the center of gravity of the characters into consideration in our subject. Suppose we consider the point $v_{ti}$, the center of gravity is the point $v_{tg}$, for the point $v_{tj}$ in the same frame, set its label to

$$l_{ti}(v_{tj}) = \begin{cases} 0 & r_i = r_j \\ 1 & r_i < r_j \\ 2 & r_i > r_j \end{cases}$$

where $r_k = d(v_{tg}, v_{tk})$, that is, the path distance from point $v_{tk}$ to the center of gravity point $v_{tg}$. That is, for neighboring points in the same frame, the distance to the center of gravity and the distance from the convolution point to the center of gravity are compared, and the points are divided into three categories according to the comparison result. The advantage of this classification is that the points far away from the center of gravity and the points close to the center of gravity can be processed separately for the human body to achieve a better feature extraction effect. After that, for neighboring points in the time dimension, define

$$l_{ti}(v_{qj}) = l_{ti}(v_{tj}) + 3(q-t)$$

That is, independent labels are assigned to the three different types of points in each frame. Finally, the convolution operation in ST-GCN can be defined as

$$f_{out}(v_{ti}) = \sum_{v_{qj} \in B(v_{ti})} \frac{1}{Z_{ti}(v_{qj})} f_{in}(v_{qj}) \cdot w(l_{ti}(v_{qj}))$$

where $Z_{ti}(v_{qj}) = \|\{v_{tk}|l_{ti}(v_{tk}) = l_{ti}(v_{tj})\}\|_1$ is the standardized item, and $w(\cdot)$ maps the label to the corresponding weight value. That is, a weighted sum of each point in the neighborhood is used as the output value of that point.

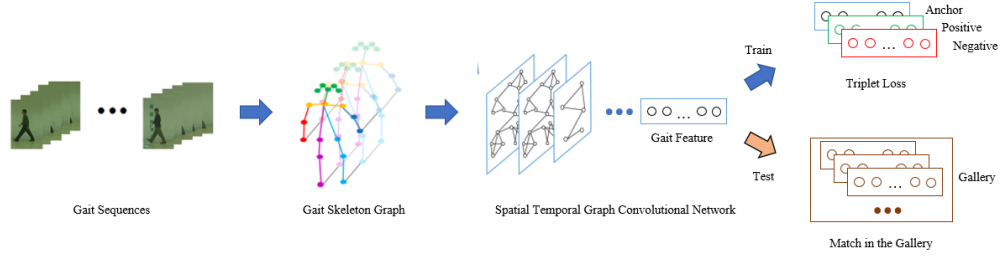

Fig. 1. the framework of the proposed system from video to the prediction

*D. Similarity-based recognition*

The similarity-based recognition is based on the following rules.
$$I_p = \{y_i | i = argmax_{i \in \{1,2,\ldots n\}} s_i\}$$
$$s_i = D(f(x_p), f(x_i))$$

The features of the sample to be tested are compared with the features of all library samples, and the label of the sample with the highest similarity is taken as the prediction result. Here $D(\cdot)$ is a method of similarity measurement. In our experiment, we use the L2 paradigm as the measurement function. Intuitively speaking, using this method means that we hope that the gait samples under the same label are as close as possible in the feature space, and the distance between the samples of different labels is as far as possible.

Since the fully connected layer is no longer needed to calculate the similarity, our model only contains ST-GCN to extract gait features. But for such a method, since the model only outputs feature vectors in the end, the biggest problem lies in the design of the loss function and the training method. If the randomly sampled batches are paired in pairs, the distance between similar samples is shortened by MSE loss, and the distance between heterogeneous samples becomes longer, then the same problem of uneven positive and negative samples as the previous method will be faced.

Finally, referring to the training method used in face recognition and pedestrian re-recognition [12], we use Triplet Loss as the loss function. Triplet Loss was originally proposed in a paper on face recognition [11]. Its goal is to train the model to learn good feature expressions, so that the trained model is not limited to a fixed category, but can extract effective features for all samples.

The basic principle is that for a set of triples $<a, p, n>$, Triplet Loss is
$$L = \max(d(a,p) - d(a,n) + margin, 0)$$
where $a$ (anchor) is an anchor point sample, $p$ (positive) is a sample of the same category as the anchor point sample, that is, a positive sample, and $n$ (negative) is a sample of a different category from the anchor point sample, that is, a negative sample. $d(\cdot)$ is a certain distance measurement method, here we use L2 distance as $d$. Therefore, Triplet Loss is, for a certain anchor point a, the difference between the distance between the positive sample and the negative sample plus a set margin. Intuitively speaking, Triplet Loss tries to narrow the distance between a and p, and widen the distance between a and n, until the distance of the former is at least a constant margin smaller than the distance of the latter. Using Triplet Loss, in an ideal situation, the final sample feature of each category clusters into a cluster, and there is a distance of at least the margin size from the clusters of other categories.

When actually using the loss function, you need to consider how to construct triples and other issues. Suppose we let the batch size be B=PK, where there are P identities, and each identity has K samples. After that, for each sample in the batch, take it as the anchor sample a, take
$$p = argmax_{p \in +} d(a,p)$$
$$n = argmin_{n \in -} d(a,n),$$

construct three Tuple $<a, p, n>$. Among them, + and - respectively represent the sample sets of the same type and different types in the batch as the anchor sample. Intuitively speaking, for each anchor point sample, we construct the hardest sample, which can also be said to be the sample most in need of optimization, that is, take the farthest positive sample and the closest negative sample. Therefore, this construction method is also called Batch Hard. Since there are a total of PK samples, construct a triple for each sample, and a total of PK triples will be obtained to calculate the loss value. After that, the PK loss values obtained are averaged to obtain the final Loss.

*E. Network implementation*

In fig. 2, the model reads a set of gait sequences as input, the batch size is 32=8*4, of which 8 people are randomly selected, and each person takes 4 samples. First, the input data is batch normalized through the BN layer, where each channel of each point is normalized in the time dimension. After that, the data will pass through the continuous ST-GCN layer for feature extraction. The ST-0 layer turns the input three-channel data into 64 channels through convolution operation. After that, ST-1, ST-2, and ST-3 are all 64-channel convolutional layers for input and output, and ST-4 inputs 64 channels. 128 channels are output, and the convolution step length is set to 2, that is, the length of the output time dimension is reduced by half to keep the number of features unchanged. After that, ST-5 and ST-6 are convolutional layers from 128 channels to 128 channels, ST-7 changes 128 channels to 256 channels, and the same convolution step size is set to 2. The last two layers ST-8 and ST-9 are 256-channel to 256-channel convolutional layers. The size of the convolution kernel of all convolutional layers is K=1 in the spatial dimension, that is, only neighbors within 1 are considered, and Γ=9 in the time dimension, that is, 4 frames before and after the current frame are considered. All convolutional layers are activated by ReLU. It should be noted that the original ST-GCN implementation introduced the Residual mechanism, but when it processes convolution with a step size of 2, it only retains the residuals for half of the data in the time dimension, resulting in each pool After transformation, half of the data lacks the gradient flow, which affects the final performance, so this part is modified in our implementation.

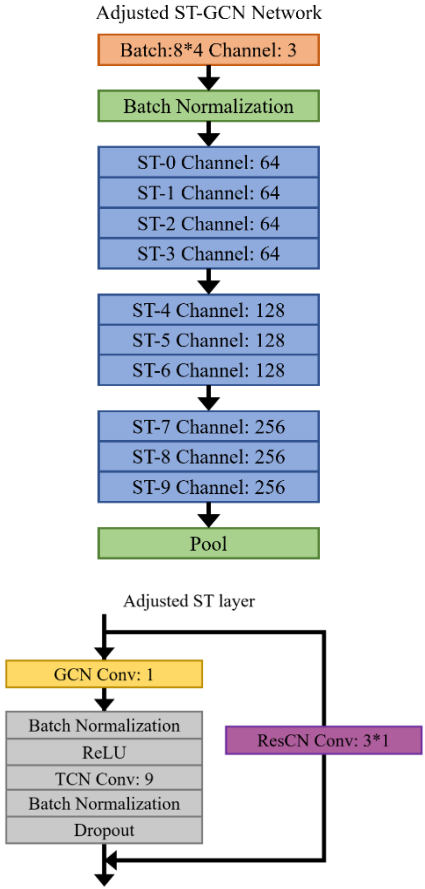

Fig. 2. the framework of Adjusted ST-GCN

Finally, in order to extract features of the same dimension for sequences of different lengths, the MaxPooling layer is added to pool the time and space dimensions, and the most significant features in each channel are extracted. The final model outputs a 256-dimensional feature vector for each sample, and calculates Triplet Loss for the 32-segment feature vector as the final optimized loss function.

## IV. EXPERIMENTS

This experiment was tested on the CASIA-B gait database whose data in the database was first preprocessed by OpenPose and extracted as a bone sequence and then used as a data set for single-view training and testing of the model.

### A. Datasets

The CASIA-B dataset contains 124 different tags. Each tag is divided into ten views, ranging from 0° to 180°, divided into a view every 18°, and the camera angles under different views are different. Each person has ten samples under each view divided into three categories: NM, BG, and CL. Among them, nm represents a normal sample, there are six segments NM #1-6, BG represents a sample carrying a bag, there are two segments BG #1-2, and CL represents a sample wearing thick clothes, and there are two segments CL #1-2.

The experimental setting is shown in Table 1. The model used the first 62 identities as the training set and the last 62 identities as the training set. During the test, the library sample was NM #1-4; the sample to be tested is divided into three parts: NM #5-6, BG #1-2, CL #1-2, and the average accuracy of the three samples to be tested was recorded in the Average column.

### B. Experimental results

Table 2 shows the comparison between our method and the three previously proposed models GaitGAN [13], SPAE [14] and PGCN [15]. It can be seen that under NM, our model performs worse than the other two models, but under BG and CL, our model shows apparent advantages. It can be seen from the average accuracy that our model has a precision that exceeds the previous method under all samples to be tested.

The main advantage of our method lies in its high adaptability to gait under different conditions. Although the previous way can achieve high accuracy on NM, the accuracy will drop significantly when the scene is changed to BG or CL. For example, GaitGAN even has an accuracy of less than 50% under CL. In our method, under most views, the accuracy of NM and BG are not much different, even at 72° and 162°. At the same time, the accuracy of CL has obvious advantages compared with other methods, and the difference between BG and NM in the view near 90° is minimal.

As SPAE and GaitGAN are based on the gait contour map to recognize the method, it can be seen that the advantages of the bone-based gait recognition method. Besides, our method performs better than PGCN, which means our method seems more suitable to detect the difference between human gaits. With the help of the previous development of the critical point estimation method, the accuracy of the bone sequence extracted from the gait video is very high, so in extraction, the influence of part of the BG and CL is automatically removed. Our model uses bone sequences as input, so compared to the contour map-based method, our model uses less information but achieves better results. The accuracy of NM is also affected by the amount of information. Visual-based methods can obtain additional details under NM to help judges, such as height, short, fat and thin, and even hairstyle and head contours. Therefore, vision-based methods can easily obtain extremely high accuracy for NM samples, but, likely, the model learned is not the gait feature. This issue needs further study. Since only bone information is used, the bone-based method does not have similar problems.

TABLE I. THE EXPERIMENTAL SETTING OF THE PROPOSED MODEL ON CASIC-B

|  | Training | Test | |
|---|---|---|---|
|  |  | Gallery set | Probe set |
| ID | 001-062 | 063-124 | 063-124 |
| Seqs | NM 01-06<br>BG  01-02<br>CL  01-02 | NM 01-04 | NM 05-06<br>BG  01-02<br>CL  01-02 |

TABLE II. THE ACCURACY OF THE PROPOSED MODEL IN SINGLE DIRECTION COMPARED WITH BASELINE

| NM #5-6 | 0° | 18° | 36° | 54° | 72° | 90° | 108° | 126° | 144° | 162° | 180° | Mean |
|---|---|---|---|---|---|---|---|---|---|---|---|---|
| GaitGAN[13] | **100** | **99.2** | 97.6 | 96 | **100** | 98.4 | 97.6 | **99.2** | **99.2** | **99.2** | **100** | **98.8** |
| SPAE[14] | 98.4 | **99.2** | 97.6 | 96 | 96 | 96 | 96.8 | 98.4 | 97.6 | 96.8 | 100 | 97.5 |
| PGCN[15] | 98.4 | **99.2** | 98.4 | **97.6** | 98.4 | 98.4 | **99.2** | **99.2** | 98.4 | **99.2** | 98.4 | 98.6 |
| Ours | 83.1 | 92.7 | **99.2** | 92.7 | 94.4 | 95.1 | 96 | 95.2 | 95.2 | 93.5 | 91.9 | 93.5 |
| BG #1-2 | 0° | 18° | 36° | 54° | 72° | 90° | 108° | 126° | 144° | 162° | 180° | Mean |
| GaitGAN[13] | 79 | 76.6 | 75.8 | 68.6 | 65.3 | 64.5 | 69.4 | 73.4 | 73.4 | 76.6 | 77.7 | 72.8 |
| SPAE[14] | 79.8 | 81.5 | 70.2 | 66.9 | 74.2 | 65.3 | 62.1 | 75.8 | 72.6 | 68.6 | 74.2 | 71.9 |
| PGCN[15] | **82.3** | 84.7 | 82.3 | 79.0 | 76.6 | 78.2 | 74.2 | 73.4 | 73.4 | 69.4 | 73.4 | 77.0 |
| Ours | 76.6 | **87.9** | **89.5** | **91.1** | **94.4** | **91.9** | **87.9** | **87.1** | **90.3** | **93.5** | **80.6** | **88.3** |
| CL #1-2 | 0° | 18° | 36° | 54° | 72° | 90° | 108° | 126° | 144° | 162° | 180° | Mean |
| GaitGAN[13] | 25.8 | 37.9 | 45.2 | 55.7 | 43.6 | 48.4 | 47.6 | 46 | 43.6 | 35.5 | 27.4 | 41.5 |
| SPAE[14] | 44.4 | 49.2 | 46.8 | 46.8 | 49.2 | 42.5 | 46.8 | 43.6 | 40.3 | 41.3 | 42.7 | 44.9 |
| PGCN[15] | 53.2 | **58.1** | 60.5 | **63.7** | 60.5 | 59.7 | 61.3 | 58.1 | 59.7 | 61.3 | 56.5 | 59.3 |
| Ours | **63.7** | 58.1 | **62.1** | 60.5 | **82.3** | **86.3** | **81.5** | **73.4** | **62.9** | **66.9** | **74.2** | **70.2** |
| Average | 0° | 18° | 36° | 54° | 72° | 90° | 108° | 126° | 144° | 162° | 180° | Mean |
| SPAE[14] | 68.3 | 76.6 | 71.5 | 69.9 | 73.1 | 67.9 | 68.6 | 72.6 | 70.2 | 68.9 | 72.3 | 71.0 |
| GaitGAN[13] | 68.3 | 71.2 | 72.9 | 73.4 | 69.6 | 70.4 | 71.5 | 72.9 | 72.1 | 70.4 | 68.4 | 71.4 |
| PGCN[15] | **78.0** | **80.7** | 80.4 | 80.1 | 78.5 | 78.8 | 78.2 | 76.9 | 77.2 | 76.6 | 76.1 | 78.3 |
| Ours | 74.5 | 79.6 | **83.6** | **81.4** | **90.4** | **91.1** | **88.5** | **85.2** | **82.8** | **84.6** | **82.2** | **84.0** |

TABLE III. COMPARISON WITH DIFFERENT SETTINGS ON 90° VIEW OF CASIA-B BY ACCURACIES

|  | Uniform | Distance | Spatial | Shallow | Deeper |
|---|---|---|---|---|---|
| Probe NM #5-6 | 0.935 | 0.976 | 0.927 | 0.927 | 0.944 |
| Probe BG #1-2 | 0.943 | 0.968 | 0.935 | 0.911 | 0.927 |
| Probe CL #1-2 | 0.895 | 0.806 | 0.895 | 0.823 | 0.895 |
| Average | 0.924 | 0.917 | 0.919 | 0.887 | 0.922 |

*C. Comparison between different settings*

To explore the influence of partition strategy and network depth on model performance, we show the identification accuracies with different network settings in Table 3.

For partition strategy, in addition to the spatial method introduced before, we test the uniform case and distance case. In uniform cases, we assign an identical label to all nodes in a single frame. In distance cases, the label is distributed according to the distance to the anchor node. The result shows all three strategies achieve similar average identification accuracies. The uniform case, which is the simplest partition strategy, presents the best performance with a margin of 0.05. It reveals the possibility that introducing a strong prior in node partition is not necessary. However, the spatial partition is used in our method because we find it is helpful to produce a stable and better performance when taking all views into consideration.

For different network depths, we test a shallow network and a deeper one. In particular, we remove three layers from each ST layer block, with 64, 128, and 256 channels, respectively, to generate a shallow model. We add two layers with 128 and 256 channels to generate a deeper model. We can figure out from the result that increasing network depth can bring better performance. But the performance enhance is reduced as the model goes deeper, with 0.032 improvements from shallow model to normal model and 0.003 improvements from normal model to deeper model. As a deeper network will bring additional cost on parameter number and convergence speed, we choose the current network setting as the final model.

The result also shows the stability of our model. Except for the shallow model with insufficient expression ability, models with other settings get approximate performance distributed in a range less than 0.01. It proves that our model can produce a stable and credible result and is not sensitive to the change of configurations.

## V. CONCLUSION

In this article, we propose a bone-based end-to-end gait recognition network. We extracted the bone sequence from the gait video through the critical point extraction method and then mapped the bone sequence to the gait diagram and connected the joint points in the spatial and temporal dimensions. We chose the method of similarity comparison to define the problem. Based on ST-GCN to extract gait features, we constructed a model and trained through Triplet Loss to get generalized gait feature expression. Tested on the CASIA-B dataset, The results of our model show that our model only uses bone information. In some cases, and on average, it has an accuracy that exceeds other methods. Compared with other methods, our approach has the advantages of solid generalization, less affected by the target state, and less information required.

There are still some problems in our current research. Future work may focus on how to change the network structure to make it more suitable for gait feature extraction and how to perform data enhancement to reduce model overfitting. In addition, the bone sequence obtained through pose estimation is not perfect, and there are often interruptions and deletions in the sequence, which significantly affects the effect of the bone-based model. After this, further research is needed to deal with or supplement the lack of bone sequence.

## VI. ACKNOWLEDGMENT

This work is funded by the Nature Natural Science Foundation of China (62002220). Ke Xu is the corresponding author.